\documentclass[10pt,twocolumn,letterpaper]{article}

\usepackage[pagenumbers]{cvpr} 

\definecolor{cvprblue}{rgb}{0.21,0.49,0.74}
\usepackage[pagebackref,breaklinks,colorlinks,allcolors=cvprblue]{hyperref}

\usepackage{amssymb}
\usepackage{amsmath}
\usepackage{color} 
\usepackage{makecell} 
\usepackage{multirow} 
\usepackage{float}
\usepackage{amsmath}
\usepackage{graphicx}
\usepackage{svg}
\usepackage{booktabs}
\usepackage{array}
\usepackage{threeparttable}
\usepackage{subcaption}

\def\paperID{*****} 
\def\confName{CVPR}
\def\confYear{2025}

\title{InterpIoU: Rethinking Bounding Box Regression with Interpolation-Based IoU Optimization}

\author{Haoyuan Liu\\
Waseda University\\
{\tt\small liuhaoyuan@akane.waseda.jp}
\and
Hiroshi Watanabe\\
Waseda University\\
{\tt\small hiroshi.watanabe@waseda.jp}
}

\begin{document}
\maketitle

\begin{abstract}
Bounding box regression (BBR) is fundamental to object detection, where the regression loss is crucial for accurate localization. Existing IoU-based losses often incorporate handcrafted geometric penalties to address IoU's non-differentiability in non-overlapping cases and enhance BBR performance. However, these penalties are sensitive to box shape, size, and distribution, often leading to suboptimal optimization for small objects and undesired behaviors such as bounding box enlargement due to misalignment with the IoU objective.
To address these limitations, we propose InterpIoU, a novel loss function that replaces handcrafted geometric penalties with a term based on the IoU between interpolated boxes and the target. By using interpolated boxes to bridge the gap between predictions and ground truth, InterpIoU provides meaningful gradients in non-overlapping cases and inherently avoids the box enlargement issue caused by misaligned penalties. Simulation results further show that IoU itself serves as an ideal regression target, while existing geometric penalties are both unnecessary and suboptimal.
Building on InterpIoU, we introduce Dynamic InterpIoU, which dynamically adjusts interpolation coefficients based on IoU values, enhancing adaptability to scenarios with diverse object distributions. Experiments on COCO, VisDrone, and PASCAL VOC show that our methods consistently outperform state-of-the-art IoU-based losses across various detection frameworks, with particularly notable improvements in small object detection, confirming their effectiveness.
\end{abstract}

\section{Introduction}
\label{sec: intro}

Object detection is a fundamental task in computer vision and plays a vital role in various real-world applications such as intelligent surveillance~\cite{ped_1_, surveillance_e1}, industrial automation~\cite{edge_e1,industrial_e1} and autonomous driving~\cite{driving_1, driving_2,vehicle_e1,vehicle_e2}. It comprises two primary sub-tasks: object classification and localization. The spatial location of an object is typically represented by a bounding box (Bbox), which encodes its position and shape. The core of localization is Bounding Box Regression (BBR), a training process that aims to maximize the geometric overlap between the predicted Bbox and its ground-truth counterpart. This overlap is universally quantified by the Intersection over Union (IoU) metric. Consequently, the design of an effective localization loss should be fundamentally aligned with the objective of maximizing the IoU value, as this directly reflects localization quality.

To facilitate effective BBR, modern object detectors—such as the RCNN family~\cite{rcnn,fast-rcnn,faster-rcnn}, SSD~\cite{ssd}, RetinaNet, DINO~\cite{dino}, and the YOLO series~\cite{yolov1,yolov3,yolov4,yolov5,yolov7,yolov8,yolox}—rely on carefully designed localization loss functions. In practice, most of these detectors adopt a combination of $\ell_n$-norm losses and IoU-based losses. However, these formulations are not fully aligned with the IoU metric. To understand the origin of this misalignment and its impact, we briefly revisit the evolution of localization loss functions in object detection.

Early object detectors such as Fast R-CNN~\cite{fast-rcnn} and YOLOv1~\cite{yolov1} employed $\ell_n$-norm losses to regress Bbox coordinates. These losses treat the four coordinates independently and ignore the geometric structure of the Bbox. As a result, they are inherently misaligned with the BBR objective and are also sensitive to object scale.
To bridge this gap, IoULoss~\cite{IoUloss} was introduced to directly optimize the IoU metric. However, it suffers from \textit{gradient vanishing} when the predicted and ground-truth boxes do not overlap, limiting its optimization effectiveness. To address this, a series of extended IoU-based losses—such as GIoU~\cite{giou}, DIoU~\cite{diou}, CIoU~\cite{ciou}, SIoU~\cite{siou}, and PIoU~\cite{piou}—propose augmenting IoU with additional geometric penalty terms (e.g., center distance, aspect ratio, or angle).
Despite their empirical improvements, most of these geometric penalties are introduced either to alleviate gradient vanishing in non-overlapping cases or to improve convergence efficiency by providing auxiliary geometric guidance. However, as they are not explicitly designed to maximize IoU, these formulations may impose optimization objectives that diverge from the fundamental goal of BBR. This misalignment can result in unintended behaviors, such as the \textit{bounding box enlargement} problem~\cite{giou}, where predicted boxes are overly expanded to minimize the auxiliary penalties rather than to genuinely increase overlap with the ground truth.
In addition, the effectiveness of these penalty terms often hinges on specific Bbox distributions. Simple geometric terms typically struggle to generalize across diverse object layouts, while more sophisticated ones introduce extra complexity with limited gain. These issues are especially problematic in scenarios involving small or crowded objects~\cite{small_e1, small_e2, small_e3, small_e4, small_e5_}, where precise and stable gradient signals are essential for accurate localization.

To address the limitations of prior IoU-based losses, we propose \textit{InterpIoU}, a novel loss function that eliminates handcrafted geometric penalties in favor of a purely IoU-driven formulation. Specifically, it introduces a penalty term defined as the IoU between an interpolated Bbox and the ground truth. By choosing a suitable interpolation coefficient, the interpolated box is guaranteed to partially overlap with the ground truth, thereby resolving the \textit{gradient vanishing} problem that occurs when predicted and ground-truth boxes do not intersect.
More importantly, since the entire loss is derived solely from IoU computations, InterpIoU preserves the desirable optimization properties of the original IoU loss and remains fully aligned with the fundamental goal of BBR—maximizing IoU. This intrinsic alignment also eliminates the risk of undesired optimization behaviors, such as the \textit{bounding box enlargement} problem introduced by auxiliary geometric penalties. 
Finally, by discarding all handcrafted geometric terms, InterpIoU yields stable and geometry-agnostic gradients, making it particularly effective in challenging scenarios involving small or densely packed objects, where conventional losses often underperform due to unstable or ineffective gradients.

To further enhance flexibility and generalization, we introduce \textit{Dynamic InterpIoU} (D-InterpIoU), which adaptively adjusts the interpolation coefficient based on the current IoU value between prediction and ground truth. This dynamic strategy creates a \textit{gradient boost zone} in low-IoU regimes, where the optimization landscape is typically flat. By strengthening the gradient signal in these regimes, D-InterpIoU enables more effective optimization for initially inaccurate predictions and leads to higher convergence accuracy. Moreover, the adaptive nature of the interpolation allows the loss function to automatically adjust to objects of varying shapes, sizes, and spatial distributions—enhancing robustness across diverse detection scenarios. Compared to static InterpIoU, the dynamic formulation provides greater generalization without introducing complexity or handcrafted components, offering a more principled and effective optimization strategy for BBR.

Extensive simulation experiments confirm that the IoU between interpolated boxes and ground truth alone serves as a strong and sufficient supervisory signal for BBR. This directly challenges the prevailing belief that auxiliary geometric penalties are necessary for effective optimization. Our results indicate that handcrafted geometric penalties are not only redundant but can even hinder performance by introducing misaligned objectives. Furthermore, benchmark evaluations on COCO, VisDrone, and PASCAL VOC show that both InterpIoU and D-InterpIoU consistently outperform previous IoU-based losses across multiple detection architectures, with particularly significant improvements in small object detection.

The main contributions of this paper are summarized as follows:
\begin{enumerate}
\item We propose InterpIoU, a novel IoU-based loss function that replaces handcrafted geometric penalties by computing IoU over interpolated boxes between the prediction and the ground truth. This design reduces the loss's sensitivity to box size, aspect ratio, and spatial distribution, and effectively addresses the issue of bounding box enlargement by maintaining better alignment with the IoU objective.

\item We further introduce Dynamic InterpIoU (D-InterpIoU), which adaptively adjusts interpolation coefficients based on the IoU value. This adaptive mechanism enhances robustness to varying object scales and spatial distributions, yielding particularly significant improvements in small object detection.

\item We further demonstrate the effectiveness of our method through simulation experiments, which highlight the unnecessary complexity of conventional geometric penalties. Additionally, our gradient analysis reveals how IoU loss design impacts the stability of BBR.

\item We integrate InterpIoU and D-InterpIoU into representative object detectors such as YOLOv8 and DINO, and validate their effectiveness on COCO, VOC07, and VisDrone benchmarks, achieving competitive detection performance among existing IoU-based losses.
\end{enumerate}


\section{Related Works}
\label{sec: related works}
\subsection{localization Loss}
Object localization is a critical sub-task of object detection. In 2-D object detection, the localization of a specific target is represented by the smallest enclosing rectangle around the target, commonly referred to as a Bbox. A Bbox encodes the target's positional information, typically in the form of its center coordinates $(x, y)$ and its size $(w, h)$. While Bboxes can be expressed in various formats, we represent them here as $B=[x,y,w,h]$.

\begin{figure}[h!tb]
    \centering
    \includegraphics[width=\linewidth]{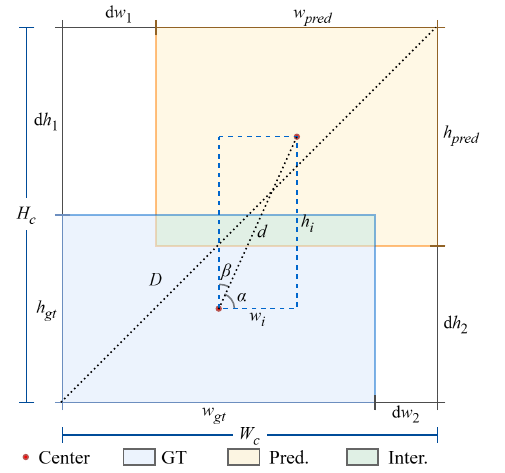}
    \caption{The geometry factors that are considered in previous IoU-based losses}
    \label{fig:geometry factors}
\end{figure}

BBR aims to refine the predicted Bbox $B_{\text{pred}}$ to closely match the ground-truth Bbox $B_{\text{gt}}$. This is typically achieved by minimizing a loss function that quantifies the discrepancy between the two boxes, formulated as:
\begin{equation}
  \text{Minimize} \quad \mathcal{L}(B_{\text{pred}}, B_{\text{gt}}),
\end{equation}
where $\mathcal{L}$ is a task-specific loss function constructed based on the chosen similarity or alignment criterion between $B_{\text{pred}}$ and $B_{\text{gt}}$.
Adopting different criteria to measure the similarities between Bboxes can construct various loss functions. 

Early approaches primarily rely on \( \ell_n \)-norm losses:
\begin{equation}
  \ell_{n}(B_{\text{pred}}, B_{\text{gt}}) = \left\|B_{\text{gt}} - B_{\text{pred}}\right\|_n,
  \label{equ:lnloss}
\end{equation}
where \( \ell_1 \) and Smooth-\( \ell_1 \) losses are commonly used in practice. YOLOv1~\cite{yolov1} adopts a square-root transformation to mitigate the influence of large box values. In YOLOX~\cite{yolox}, \( \ell_1 \) loss is introduced as an auxiliary loss in the final training epochs to enhance BBR performance. Studies have shown that \( \ell_n \)-norm losses are effective in improving localization accuracy. 

Despite their effectiveness, \( \ell_n \)-norm losses treat the bounding box parameters (e.g., center coordinates, width, and height) as independent scalar variables, and thus fail to capture the spatial and geometric correlations between boxes. 

To better quantify the similarity between Bboxes, Intersection over Union (IoU) has been proposed as a classical metric based on the areas of overlap and union between Bboxes. It is defined as:
\begin{equation}
  \text{IoU}(B_{\text{pred}}, B_{\text{gt}}) = \frac{B_{\text{pred}} \cap B_{\text{gt}}}{B_{\text{pred}} \cup B_{\text{gt}}}.
  \label{equ:IoU}
\end{equation}

Therefore, the limitations of the $\ell_n-norm$ loss motivate the development of IoU-based loss, which offers a more holistic view towards the BBR, thus more aligned with the IoU evaluation metrics.

The original IoU loss directly employs the negative IoU value as the optimization target, leveraging the fact that IoU increases as the predicted box better aligns with the ground truth. In \cite{IoUloss}, which first applied the IoU-based loss, the authors used a logarithmic transformation to the IoU score for a better BBR. Subsequent studies have commonly adopted a simplified formulation using the negative IoU value, as follows:
\begin{equation}
  \mathcal{L}_\text{IoU}(B_{\text{pred}}, B_{\text{gt}}) = 1- \text{IoU}(B_{\text{pred}}, B_{\text{gt}}),
  \label{equ:IoU loss}
\end{equation}which captures the geometric consistency between bounding boxes and aligns more closely with evaluation metrics used in object detection.

\subsection{The Extension of the IoU-Based Loss}
The $\mathcal{L}_{IoU}$ (Eq. \ref{equ:IoU loss}) has proven effective for BBR when combined with traditional $\ell_n$-norm losses. However, it cannot be used independently due to a fundamental limitation: its gradient vanishes when the predicted and ground-truth boxes do not overlap. To overcome this issue, various penalty terms have been proposed and integrated with the IoU loss—either additively or multiplicatively—leading to a generalized formulation:
\begin{equation}
    \mathcal{L}' = R_m \cdot \mathcal{L}_{\text{IoU}} + \sum_{i} R_{a,i},
\end{equation}
where \( \mathcal{L}' \) denotes the modified IoU-based loss, and \( R_m \), \( R_{a,i} \) are additional penalty coefficients and components.

These penalty terms leverage more geometric factors to construct more effective IoU-based losses, extending the applicability of IoU loss and providing more informative gradients for diverse Bbox distributions. We briefly introduce several representative IoU-based penalties that have been proposed to enhance the IoU loss, with their geometric factors illustrated in Fig.~\ref{fig:geometry factors}.

\textbf{GIoU} \: The first extension of the IoU loss is the Generalized IoU (GIoU)~\cite{giou}, which introduces an additive penalty based on the smallest enclosing box that contains both the predicted and ground-truth boxes. This term allows gradients to be propagated even in non-overlapping cases:
\begin{equation}
    R_{\text{GIoU}} = \frac{W_c H_c - U}{W_c H_c},
    \label{eq:giou}
\end{equation}
where \( W_c \) and \( H_c \) denote the width and height of the enclosing box , and \( U \) is the union area. 

However, since the enclosing box can be significantly larger than the individual bounding boxes, this penalty may unintentionally encourage expansion of the predicted box, leading to the \textit{bounding box enlargement} issue.

\textbf{DIoU} \: Following GIoU, Distance-IoU (DIoU)\cite{diou} was proposed to incorporate more refined geometric factors. DIoU introduces a penalty term based on the normalized squared Euclidean distance between the center points of the predicted and ground-truth boxes:
\begin{equation}
    R_{\text{DIoU}} = -\frac{d^2}{D^2},
    \label{eq: diou}
\end{equation}
where $d$ and $D$ denotes the distance of the centers of the Bboxes and the the diagonal length of the minimum enclosing box. However, DIoU does not account for the shape discrepancy between the Bboxes.

\textbf{CIoU} \: Complete-IoU (CIoU)~\cite{ciou} extends DIoU by incorporating aspect ratio consistency into the penalty formulation. The overall penalty term is defined as:
\begin{equation}
    R_{\text{CIoU}} = R_{\text{DIoU}} + \alpha v,
\end{equation}
where \( v = \frac{4}{\pi^2} \left( \arctan \frac{w_{\text{gt}}}{h_{\text{gt}}} - \arctan \frac{w_{\text{pred}}}{h_{\text{pred}}} \right)^2 \) measures the difference in aspect ratios, and \( \alpha = {v}/{(1 - \text{IoU}) + v} \) is a dynamic weighting factor. When the IoU is low (i.e., $L_{\text{IoU}} < 0.5$), $\alpha$ is typically set to zero to reduce the influence of the aspect ratio term. However, introducing w/h into penalty terms results in CIoU being overly sensitive to small changes in width and height, especially in cases of thin or elongated bounding boxes.

\textbf{EIoU} \: Efficient-IoU (EIoU)~\cite{eiou} further improves upon CIoU by decoupling the width and height discrepancies into separate penalty terms. Instead of relying solely on aspect ratio similarity as in CIoU, EIoU directly minimizes the differences between the predicted and ground-truth box dimensions. The penalty term is formulated as:
\begin{equation}
    R_{\text{EIoU}} = \frac{(w_{\text{pred}} - w_{\text{gt}})^2}{W_c^2} + \frac{(h_{\text{pred}} - h_{\text{gt}})^2}{H_c^2},
    \label{eq:eiou}
\end{equation}
where \( w_{\text{pred}} \) and \( h_{\text{pred}} \) denote the width and height of the predicted box, \( w_{\text{gt}} \) and \( h_{\text{gt}} \) are those of the ground-truth box. However, its reliance on absolute size differences makes it sensitive to scale, and less effective when the boxes differ significantly in both size and position.

\textbf{WIoU} \: Wise-IoU (WIoU)~\cite{wiou}, which rethinks the design of IoU-based losses by introducing a dynamic focusing mechanism. WIoU adopts a multiplicative penalty term defined as \( R_{\text{WIoU}} = \exp\left({d^2}/{D^2}\right) \). This formulation emphasizes gradients for well-localized predictions while suppressing the influence of outliers, leading to more stable and accurate regression.

\textbf{SIoU} \: Scylla-IoU (SIoU)~\cite{siou} introduces a highly sophisticated IoU-based loss function by incorporating multiple geometric factors, including angle, distance, and shape. The penalty term is defined as:
\begin{equation}
    R_{\text{SIoU}} = \frac{\Delta + \Omega}{2}.
\end{equation}
The distance cost $\Delta$ is computed as $\Delta = 2 - e^{-\gamma \rho_x} - e^{-\gamma \rho_y}$, where $\rho_x = \frac{w_i}{W_c}$, $\rho_y = \frac{h_i}{H_c}$, and $\gamma = 2 - \Lambda$. The angle cost $\Lambda$ is defined as $\Lambda = 1 - 2\sin^2(x - \frac{\pi}{4})$, with $x = \text{min}(\alpha, \beta)$, as illustrated in Fig.~\ref{fig:geometry factors}. The shape cost $\Omega$ is given by $\Omega = (1 - e^{-\omega_w})^{\theta} + (1 - e^{-\omega_h})^{\theta}$, where $\theta \in [2, 4]$, $\omega_w = \frac{\|w_{\text{pred}} - w_{\text{gt}}\|}{\max(w_{\text{pred}}, w_{\text{gt}})}$, and $\omega_h = \frac{\|h_{\text{pred}} - h_{\text{gt}}\|}{\max(h_{\text{pred}}, h_{\text{gt}})}$.

SIoU is the first IoU-based loss to introduce angle cost into BBR, addressing challenges related to angular alignment, spatial proximity, and shape variation through a comprehensive design based on geometric analysis.

\textbf{PIoU} \: Powerful IoU (PIoU)~\cite{piou} is a recent addition to the IoU-based loss family, designed to be lighter yet more efficient compared to SIoU. PIoU introduces a target-adaptive penalty that guides regression along a more direct and effective path. The penalty term is defined as:
\begin{equation}
    R_{\text{PIoU}} = 1 - e^{-p^2},
\end{equation}
where $p = \left(\frac{\text{d}w_1}{w_{\text{gt}}} + \frac{\text{d}w_2}{w_{\text{gt}}} + \frac{\text{d}h_1}{h_{\text{gt}}} + \frac{\text{d}h_2}{h_{\text{gt}}} \right)/4$. 
However, PIoU is vulnerable to large shape differences between Bboxes, which can hinder stable optimization.

\section{Methodology}
\label{sec:proposed_method}

While incorporating geometric factors into BBR may seem intuitive, introducing additional geometric terms into the loss function is not always beneficial. On one hand, such terms may result in a loss function that no longer aligns well with the standard IoU metric used during evaluation. It remains unclear whether increasingly complex penalty terms—derived from over-engineered geometric formulations—are necessary or even effective. On the other hand, improperly designed loss components may hinder the optimization process of BBR, as exemplified by issues such as the bounding box enlargement problem.

In this section, we introduce a novel IoU-based loss function that avoids auxiliary geometric components and instead relies solely on the IoU metric. Our gradient analysis shows that the original IoU loss possesses desirable gradient flow properties that are beneficial for BBR. Furthermore, simulation experiments demonstrate that this simplified approach is both efficient and effective. These results suggest that additional geometric penalty terms are unnecessary and may even be counterproductive.

\begin{figure}[h!tb]
  \centering
  \includegraphics[width=\linewidth]{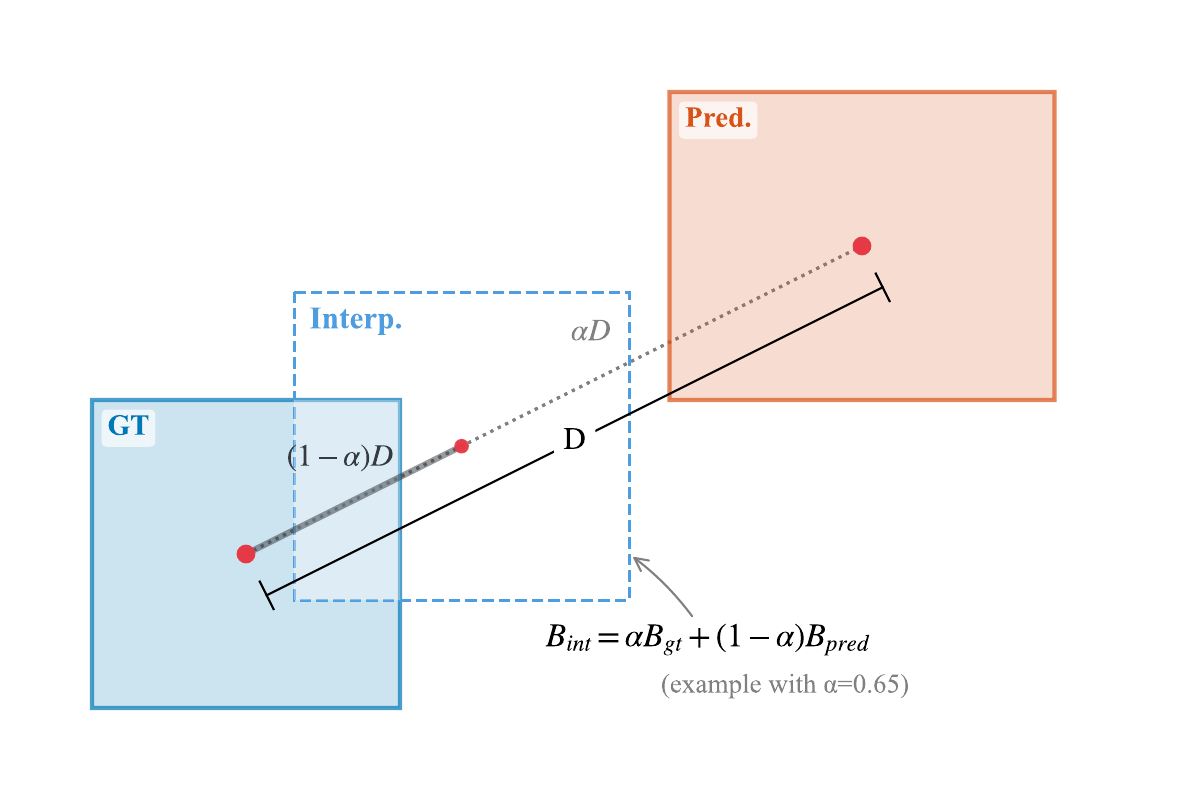}
  \caption{
    Illustration of the interpolation strategy in InterpIoU. The interpolated box ($B_{\text{int}}$) is generated as a weighted average of the Ground Truth ($B_{\text{gt}}$) and Prediction ($B_{\text{pred}}$) boxes, controlled by the coefficient $\alpha$. By positioning $B_{\text{int}}$ at a proportional distance of $(1-\alpha)D$ from the GT center, where $D$ is the total distance to the Pred center, this strategy ensures a meaningful (i.e., non-zero) IoU between the interpolated box and the ground truth, thus providing a stable regression target.
  }
  \label{fig:interp_illus}
\end{figure}

\subsection{Interp-IoU Loss}
\label{sec: interpiou proposal}

A well-known limitation of the standard IoU loss is its inability to provide meaningful gradients when the predicted and ground-truth bounding boxes do not overlap. To mitigate this issue, previous works have introduced auxiliary geometric penalty terms—such as aspect ratio constraints or distance metrics—to guide optimization. However, such handcrafted terms are often sensitive to Bbox shapes and distributions, which can lead to unstable training dynamics and degraded performance.

\textbf{Interpolated Bbox Construction} \quad To address this issue while preserving the core intuition of IoU, we propose a simple yet effective solution: an interpolated Bbox constructed between the prediction and the ground truth. This interpolated box ensures a non-zero IoU with the ground truth, thereby maintaining gradient flow even in cases of complete misalignment between the predicted and ground-truth Bboxes.

Given an interpolation coefficient $\alpha$, the interpolated Bbox $B_{int}$ is defined as:
\begin{equation}
B_{int} = (1-\alpha)B_{\text{pred}} + \alpha B_{\text{gt}}, \quad 0 < \alpha < 1,
\label{eq:interp_bbox}
\end{equation}
where $B_{\text{pred}}$ and $B_{\text{gt}}$ denote the predicted and ground-truth bounding boxes, respectively. This formulation linearly interpolates their coordinates. A larger $\alpha$ moves the interpolated box closer to the ground truth, while a smaller $\alpha$ keeps it closer to the prediction as shown in Fig.\ref{fig:interp_illus}.

We define the Interp-IoU loss as:
\begin{equation}
\mathcal{L}_{\text{InterpIoU}}(B_{\text{pred}}, B_{\text{gt}}) = \mathcal{L}_{\text{IoU}}(B_{\text{pred}}, B_{\text{gt}}) + \mathcal{L}_{\text{IoU}}(B_{int}, B_{\text{gt}}).
\label{eq:interp_iou_loss}
\end{equation}
The first term is the standard IoU loss, and the second term penalizes the divergence between the ground truth Bbox and the interpolated Bbox, which lies closer to the ground truth.

A key insight lies in the behavior of the interpolated box $B_{\text{int}}$. When $\alpha$ is sufficiently large, $B_{\text{int}}$ tends to overlap with $B_{\text{gt}}$, even when $B_{\text{pred}}$ does not. This ensures that the interpolation-based IoU is non-zero, thereby providing a useful and smooth gradient back propagation. Additionally, by penalizing the difference between $B_{\text{pred}}$ and $B_{\text{int}}$ close to $B_{\text{gt}}$, the optimization process is naturally guided toward better alignment with the target box. These interpolated boxes serve as soft geometric anchors that promote stable convergence, especially for small, sparse, or difficult-to-localize objects.

\begin{figure}[h!tb]
  \centering 
  \includegraphics[width=0.9\linewidth]{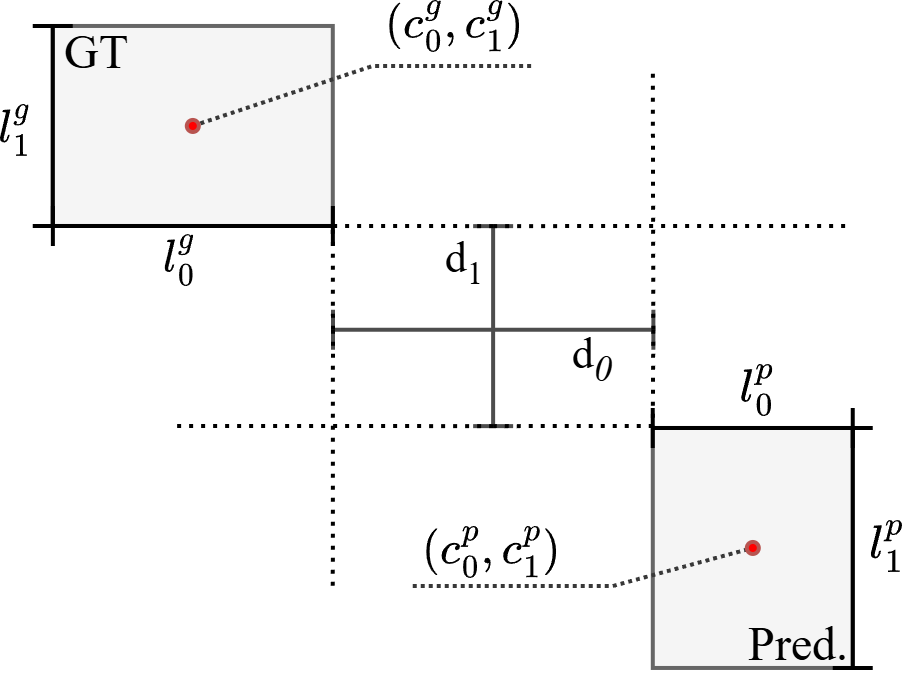}
  \caption{Illustration of the relationship between the predicted and ground-truth Bboxes along the horizontal ($i = 0$) and vertical ($i = 1$) axes.}
  \label{fig:alpha_range}
\end{figure}

\textbf{Selection of the $\alpha$} \quad We now analyze the feasible range of the interpolation coefficient $\alpha$. Without loss of generality, we consider a simple case where the predicted Bbox is located at the bottom-right corner of the ground-truth Bbox, as illustrated in Fig.~\ref{fig:alpha_range}.  
Let $c_i^g$ and $l_i^g$ denote the center coordinate and side length of the $B_{\text{g}}$ along axis $i$ ($i = 0$ for the horizontal axis, $i = 1$ for the vertical axis), and let $c_i^p$ and $l_i^p$ denote the corresponding values for the $B_{\text{p}}$. The interpolated Bbox $B'$ is defined by its center $c'_i$ and side length $l'_i$.

To ensure that the interpolated Bbox has a non-zero intersection with the ground-truth Bbox, the following condition must be satisfied:
\begin{equation}
    c_i^g + \frac{l_i^g}{2} > c'_i - \frac{l'_i}{2}, \quad i \in \{0, 1\}.
    \label{eq:alpha_range_1}
\end{equation}

We define the minimum edge-to-edge distance between the two Bboxes along axis $i$ as:
\begin{equation}
    d_i = c_i^p - c_i^g - \frac{l_i^p + l_i^g}{2}, \quad i \in \{0, 1\}.
    \label{eq:alpha_range_di}
\end{equation}

Given the linear interpolation formulation:
\begin{equation}
    c'_i = \alpha c_i^g + (1 - \alpha)c_i^p,\quad
    l'_i = \alpha l_i^g + (1 - \alpha)l_i^p,\quad i \in \{0, 1\}.
    \label{eq:interp_bbox_sec3}
\end{equation}
we can substitute Eq.(\ref{eq:interp_bbox_sec3}) into Eq.(\ref{eq:alpha_range_1}) and derive a lower bound for $\alpha$:
\begin{equation}
    \max\left\{ \frac{d_i}{d_i + l_i^g},\, 0 \,\middle|\, i = 0, 1 \right\} < \alpha < 1.
    \label{eq:alpha_final}
\end{equation}
Meanwhile, since the interpolated Bbox is defined as a convex combination of the ground-truth and predicted Bboxes, the upper bound of $\alpha$ is trivially $1$.  

This inequality guarantees that the interpolated Bbox maintains overlap with the ground-truth Bbox along both axes.

Considering the so-called major cases~\cite{wiou} and widely used label assigners such as ATSS~\cite{atss} and TAL~\cite{tal}, the initially assigned predicted Bbox and target Bbox typically satisfy the empirical condition $d_i < l_i^g$. Based on this observation, a lower bound of $\alpha > 0.5$ is sufficient to ensure intersection in most regression scenarios. As stated in the ablation study in Sec.~\ref{sec: abla on alpha}, we set $\alpha = 0.98$ in all experiments unless otherwise specified.

\textbf{Gradient Flow and Optimization Behavior} \quad From a backpropagation perspective, this loss formulation allows gradient flow through both terms:

\begin{align}
    \frac{\partial \mathcal{L}_{\text{InterpIoU}}}{\partial B_{\text{pred}}} 
    &= \frac{\partial \mathcal{L}_{\text{IoU}}^{(\text{pred}, \text{gt})}}{\partial B_{\text{pred}}}
    + \frac{\partial \mathcal{L}_{\text{IoU}}^{(\text{int}, \text{gt})}}{\partial B_{\text{int}}}
    \cdot \frac{\partial B_{\text{int}}}{\partial B_{\text{pred}}} \\
    &= \frac{\partial \mathcal{L}_{\text{IoU}}^{(\text{pred}, \text{gt})}}{\partial B_{\text{pred}}}
    + \frac{\partial \mathcal{L}_{\text{IoU}}^{(\text{int}, \text{gt})}}{\partial B_{\text{int}}} \cdot (1 - \alpha)
\end{align}

Here, \( \mathcal{L}_{\text{IoU}}^{(\text{pred}, \text{gt})} \) denotes the IoU loss computed between the predicted box and ground truth, while \( \mathcal{L}_{\text{IoU}}^{(\text{int}, \text{gt})} \) is computed between the interpolated box and the ground truth.

This derivation highlights that the second term contributes a scaled version of the gradient with respect to the interpolated box, weighted by $(1 - \alpha)$. As a result, the gradient remains well-defined and informative even in cases where the original IoU loss provides no gradient signal. 

Furthermore, since $\mathcal{L}_{\text{IoU}} \in [0,1]$, minimizing $\mathcal{L}_{\text{InterpIoU}}$ inherently drives the minimization of $\mathcal{L}_{\text{IoU}}$. This ensures that the proposed loss remains faithful to the objective of improving BBR accuracy.

In summary, the proposed Interp-IoU loss maintains the simplicity and metric consistency of the standard IoU, while overcoming its gradient-vanishing problem. Unlike prior works that depend on geometry-specific heuristics, our method is robust to diverse object shapes and distributions, and is naturally compatible with IoU-based evaluation protocols used in object detection benchmarks.

\begin{figure}[h!tb]
    \centering 
    \includegraphics[width=\linewidth]{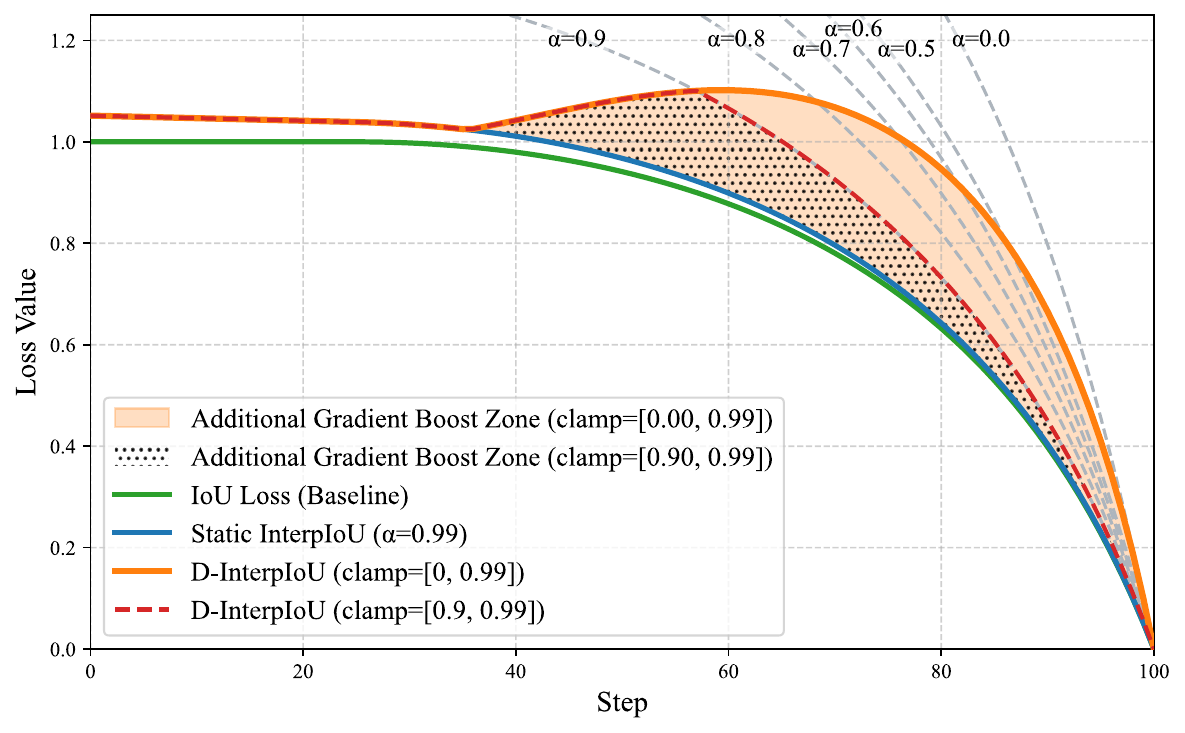}
    \caption{
     The plot visualizes the loss value as a prediction box moves towards the ground truth (from step 0 to 100).
     Our proposed D-InterpIoU with \texttt{clamp=[0, 0.99]} (orange curve) consistently forms an upper envelope over the Static InterpIoU with a fixed $\alpha=0.99$ (blue curve). 
     The shaded region between them highlights the additional gradient boost provided by our dynamic strategy.
     The behavior of D-InterpIoU is further illustrated with a stricter clamp range, \texttt{[0.9, 0.99]} (red dashed line), which follows the envelope and then converges to the static curve.
     For reference, the baseline IoU Loss and several Static InterpIoU variants with different fixed $\alpha$ values are also shown.
    }
    \label{fig:loss_landscape}
\end{figure}

\begin{figure*}[h!tb]
  \centering 
  \includegraphics[width=\linewidth]{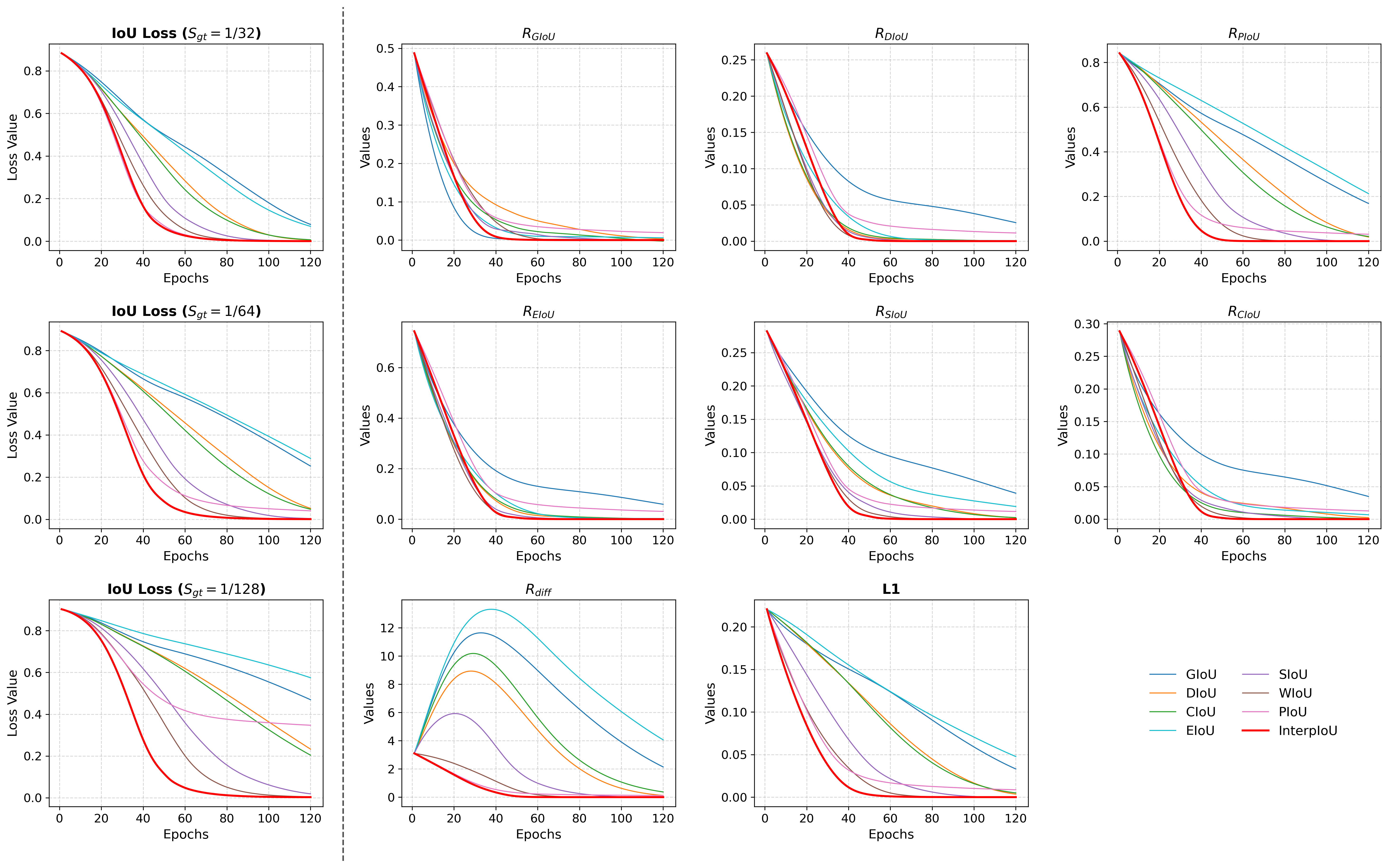}
  \caption{Visualization of the optimization behaviors of various IoU-based losses during regression. Each subfigure reports the average values over all the BBR of each steps using different IoU-based loss. The left column shows the basic IoU loss values at different target sizes. The right part of the figure reports the value changes of different geometry factors based penalty terms, including the common IoU-based loss and L1. In all the cases, the InterpIoU shows superior optimization dynamics, demonstrating that relying solely on IoU is more effective than using additional geometric penalties.}
  \label{fig:loss-performance-loss-cirteria}
\end{figure*}

\subsection{Dynamic InterpIoU: IoU-Guided Adaptive Interpolation}

While InterpIoU with a fixed interpolation coefficient $\alpha$ solves the vanishing gradient problem for non-overlapping boxes, a static setting may not be ideal across different training stages or object distributions. In particular, under a high $\alpha$, the interpolated box $B_{\text{int}}$ remains close to the ground truth even when predictions are already accurate, effectively reducing the loss to standard IoU. This limits its ability to refine Bbox during late-stage optimization. To address this, we propose \textit{Dynamic InterpIoU} (D-InterpIoU), which adapts $\alpha$ based on the current IoU, maintaining stronger learning signals throughout training.

Specifically, the dynamic interpolation coefficient $\alpha_{\text{dyn}}$ is defined as:
\begin{equation}
    \alpha_{\text{dyn}} = \text{clamp}(1 - \text{IoU}(B_{\text{pred}}, B_{\text{gt}}), \alpha_{\text{low}}, \alpha_{\text{high}}).
    \label{eq:dynamic_alpha}
\end{equation}
Here, the coefficient is inversely related to prediction quality. The clamping bounds $\alpha_{\text{low}}$ and $\alpha_{\text{high}}$ ensure numerical stability and gradient continuity. The upper bound $\alpha_{\text{high}}$ (e.g., 0.99) ensures sufficient interpolation even when $\text{IoU} = 0$, providing persistent gradients. The lower bound $\alpha_{\text{low}}$ (e.g., 0.5) ensures that the interpolated box retains a minimum degree of supervision, preventing vanishing influence of InterpIoU when predictions are nearly correct.

A distinctive feature of D-InterpIoU lies in its effect on the optimization dynamics during early training. When predictions are poor ($\text{IoU} \approx 0$), $\alpha_{\text{dyn}}$ remains near its upper limit, placing $B_{\text{int}}$ close to $B_{\text{gt}}$. As training progresses and IoU improves, $\alpha_{\text{dyn}}$ decreases, moving $B_{\text{int}}$ away from the ground truth. This can temporarily increase the interpolation loss $L_{\text{interp}}$, creating a region of \textbf{additional gradient boost}(Fig.\ref{fig:loss_landscape}). Rather than being a drawback, this transient increase provides a beneficial optimization signal: it acts as a strong driving force that helps the model escape poor local minima and flat loss regions. This adaptive behavior resembles implicit curriculum learning and exemplifies a form of dynamic loss shaping tailored to prediction difficulty.

\subsection{Simulation Experiments}  
\label{sec: simulation experiments}
To evaluate the effectiveness of InterpIoU in terms of convergence speed and eliminating the need for redundant geometric penalties, we conducted regression simulation experiments following the setups used in prior studies~\cite{ciou,wiou,piou}. The goal is to preliminarily compare different BBR loss functions in a controlled environment. We used Adam as the optimizer with a learning rate of 0.01.

In our simulation, we systematically analyzed the influence of distance, scale, and aspect ratio on regression performance. Specifically, we defined seven ground-truth boxes, all having a unit area $S_{gt}$ but with varying aspect ratios: \( 1\!:\!4, 1\!:\!3, 1\!:\!2, 1\!:\!1, 2\!:\!1, 3\!:\!1, \) and \( 4\!:\!1 \). The centers of all ground-truth boxes were fixed at \( (0.5, 0.5) \).

We generated 5,000 anchor points uniformly distributed within a circular region of radius 0.5 centered at \( (0.5, 0.5) \). At each anchor point, we assigned 7 different scales \( (1/32, 1/24, 3/64, 1/16, 1/12, 3/32, 1/8) \), along with the 7 aspect ratios mentioned above, resulting in \( 5,000 \times 7 \times 7 = 245,000 \) BBR cases in total.

\begin{figure}
    \centering
    \includegraphics[width=0.75\linewidth]{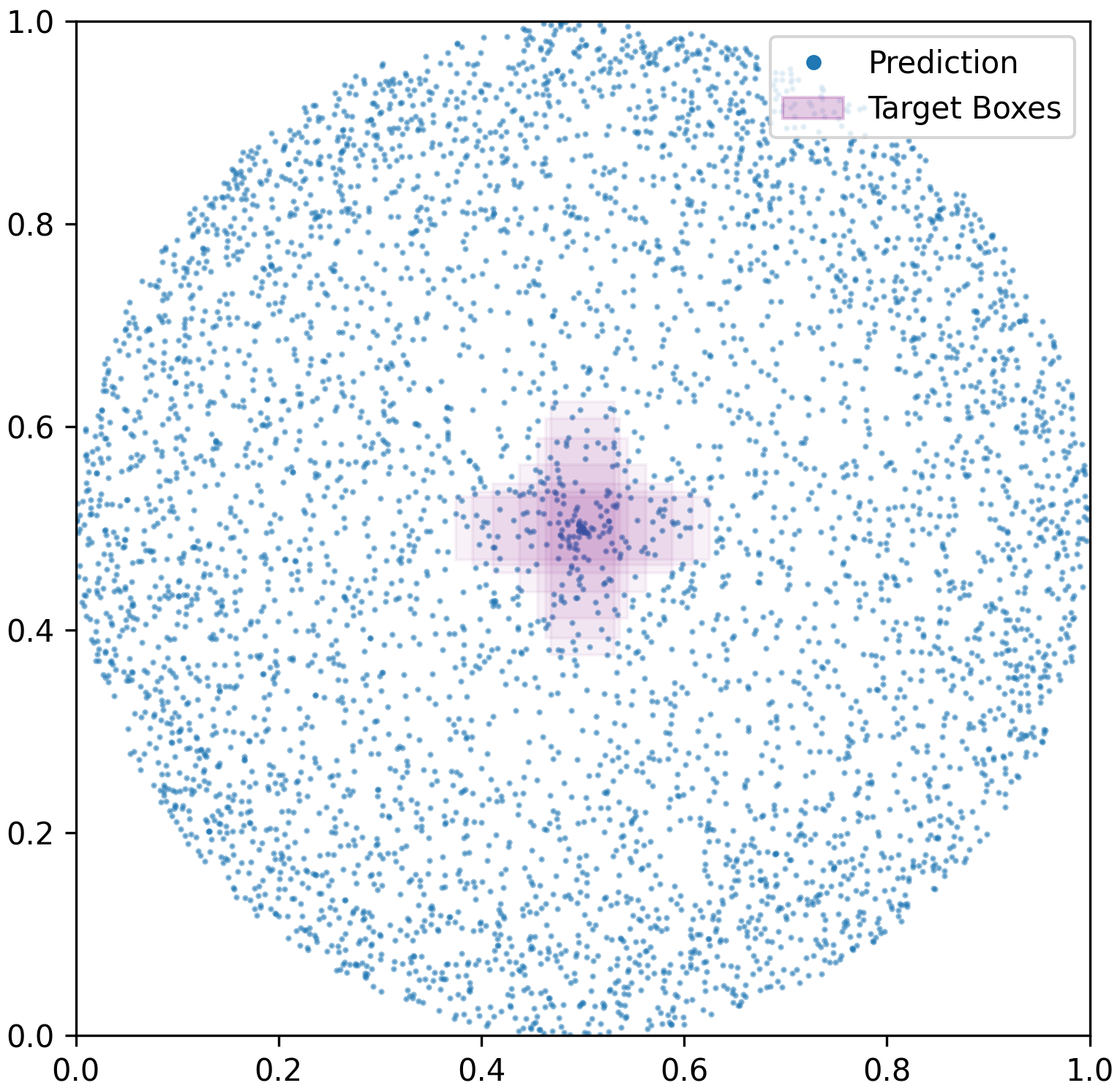}
    \caption{Visualization of the simulated prediction distribution (blue dots) and the corresponding target boxes (purple rectangles) for evaluating IoU-based loss functions. The predictions are uniformly scattered within a circular region centered on the ground-truth.}
    \label{fig:simulation illus}
\end{figure}

An illustration of the simulated prediction distribution is provided in Fig.~\ref{fig:simulation illus}, where blue dots represent the initial prediction points and the purple rectangles denote the target boxes. 

During the simulation, we recorded the designated intermediate states (e.g. IoU loss, $\ell_1$) of BBR cases at each step, and computed the average values across all cases to analyze the overall optimization behavior. Specifically, we tracked the values of each penalty term (e.g. $R_{GIoU}$ in Eq.\ref{eq:giou}) associated with different IoU-based losses with $S_{gt}=1/64$, aiming to assess their optimization behaviors under various criteria.

The results are illustrated in Fig.~\ref{fig:loss-performance-loss-cirteria}. It can be observed that InterpIoU consistently achieves superior performance across different evaluation metrics. This suggests that although many previous studies proposed various penalty terms based on geometric factors to enhance IoU-based losses, these handcrafted designs are not necessarily optimal. In fact, our results indicate that such additional geometric penalties are often redundant. The penalty terms relying solely on IoU are better.

\subsection{Solving the Bbox Enlargement Problem}

In this section, we explain how InterpIoU addresses the \textit{Bbox enlargement problem}, both through theoretical analysis and simulation experiments.

\subsubsection{Visualization of Bbox Enlargement}
Our simulation experiments reveal that certain IoU-based loss functions tend to generate excessively large predicted bounding boxes during optimization, a phenomenon previously noted in the ~\cite{wiou, piou}. This issue is particularly pronounced when inappropriate geometric factors are incorporated into the penalty terms. We refer to this behavior as the \textit{Bounding Box Enlargement Problem}, which can compromise localization accuracy and negatively impact detection performance in real-world scenarios.

To illustrate this effect, we designed a simple controlled regression experiment. The center of the initial Bbox was set at \([0.5, 0.5]\) with an aspect ratio of \(1\!:\!3\), while the target box was located at the origin \((0, 0)\) with an aspect ratio of \(3\!:\!1\). Both boxes were set to have an area of \(1/4\). Typically, the initial anchor box has no intersection with the ground truth box. We used the same optimizer and hyperparameters as in the previous simulation experiments.

\begin{figure}[h!tb]
  \centering 
  \includegraphics[width=\linewidth]{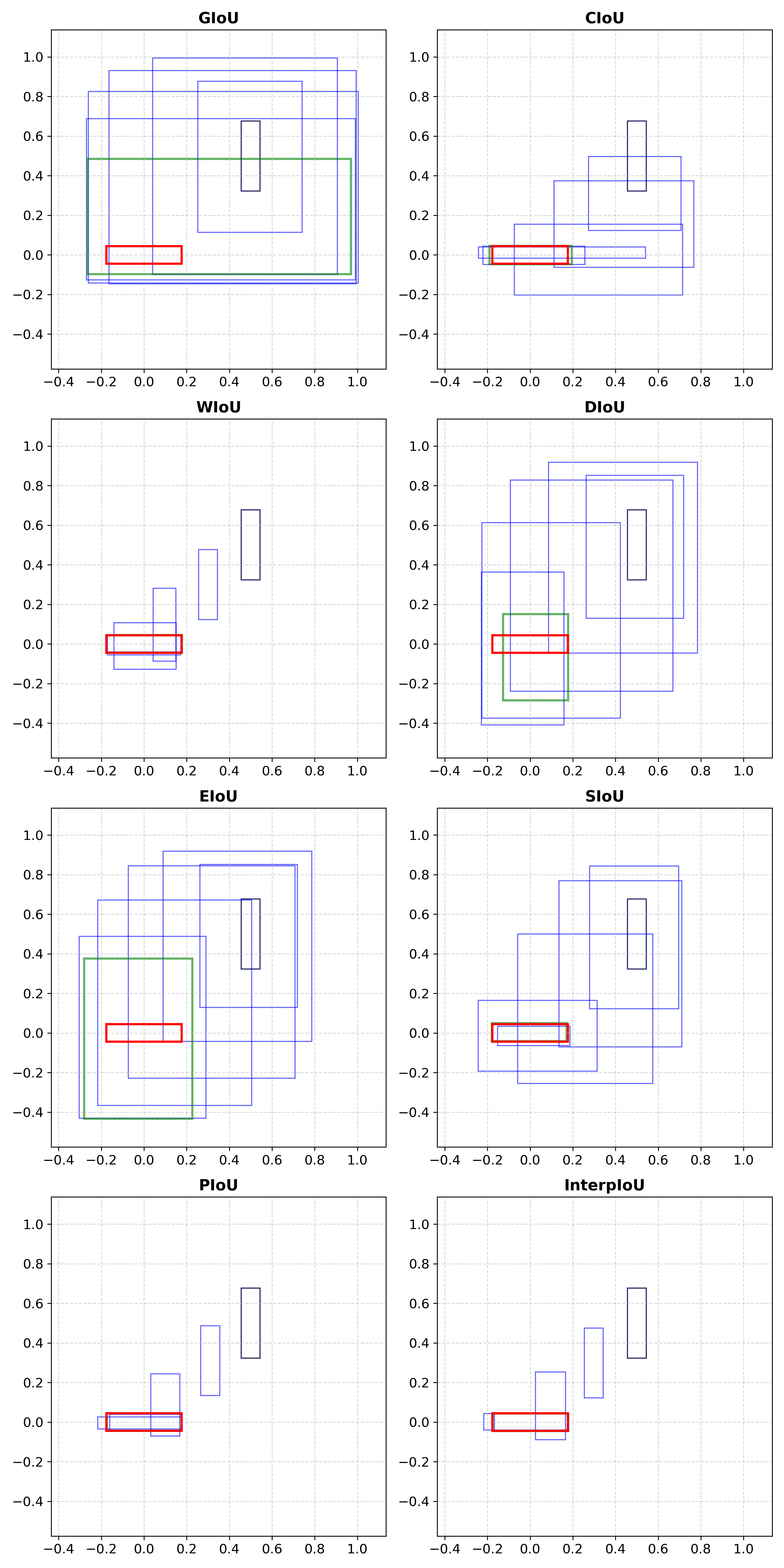}
  \caption{Visualization of BBR using various IoU-based losses. The black box indicates the initial anchor, the red box represents the target, blue boxes show the intermediate states during optimization, and the green box is the final regression result.}
  \label{fig:iou-bbox-regression}
\end{figure}

We then applied different IoU-based loss functions for regression. The visualization results are shown in Fig.~\ref{fig:iou-bbox-regression}. It is clearly observed that GIoU~\cite{giou}, CIoU~\cite{ciou}, DIoU~\cite{diou}, EIoU~\cite{eiou}, and SIoU~\cite{siou} all result in significant bbox enlargement in BBR.

To quantitatively analyze the extent of Bbox enlargement, we introduce a simple metric $R_{diff}$  to evaluate area changes during BBR:
\begin{equation}
    R_{diff} = \frac{S - S_g}{S_g},
    \label{equ: area diff}
\end{equation}
where \( S \) denotes the area of the predicted Bbox and \( S_g \) denotes that of the ground truth. A positive \( R_{diff} \) value indicates the predicted box is larger than the ground truth, while a negative value indicates it is smaller.

We computed \( R_{diff} \) for different IoU-based loss functions during the simulation experiment in Sec.\ref{sec: simulation experiments}. The results are shown in Fig.~\ref{fig:loss-performance-loss-cirteria}. As illustrated, traditional IoU-based losses such as GIoU, EIoU, DIoU, SIoU, and CIoU exhibit significant bounding box enlargement, with \( R_{diff} \) values exceeding 5 during regression. Such substantial area increases can negatively impact localization accuracy, especially in dense object detection scenarios.

In contrast, methods like WIoU, PIoU, and our proposed InterpIoU maintain \( R_{diff} \) values close to zero, indicating minimal area variation throughout the regression process. This demonstrates that InterpIoU effectively mitigates the bounding box enlargement problem, ensuring more stable and accurate optimization.

Now, we have empirically demonstrated that InterpIoU is not likely to suffer from the Bbox enlargement issue. To further unveil its internal mechanism, we deeply analyze its geometric properties and optimization behavior.

\subsubsection{Gradient Analysis}
We use $S$ and $S_g$ to denote the areas of the prediction Bbox and ground truth Bbox respectively, which can be obtained by $S_g = w_gh_g$ and $S = wh$. We then use $U$ to represent the union area of the two Bboxes: $U = S+S_g-I$, where $I$ is the area of the intersection part of the two Bboxes. Next, the basic IoU loss $L{IoU}$ is written as the 
\begin{equation}
  L_{IoU} = 1 - \frac{I}{U}. 
  \label{eq:IoU loss in analysis}
\end{equation}

We now can illustrated its effectiveness by computing the gradient of this loss w.r.t $x,y,w,h$:
\begin{align}
  \frac{\partial L_{IoU}}{\partial x} &= -\frac{S_g+S}{U^2}\cdot \frac{\partial I}{\partial x}, \label{eq:partial derivatives x of IoU} \\
  \frac{\partial L_{IoU}}{\partial y} &= -\frac{S_g+S}{U^2}\cdot \frac{\partial I}{\partial y} ,\label{eq:partial derivatives y of IoU} \\
  \frac{\partial L_{IoU}}{\partial w} &= -\frac{S_g+S}{U^2}\cdot \frac{\partial I}{\partial w} + \frac{I}{U^2}\cdot \frac{\partial S}{\partial w},  \label{eq:partial derivatives w of IoU} \\
  \frac{\partial L_{IoU}}{\partial h} &= -\frac{S_g+S}{U^2}\cdot \frac{\partial I}{\partial h} + \frac{I}{U^2}\cdot \frac{\partial S}{\partial h}.
  \label{eq:partial derivatives h of IoU}
\end{align}

From Eq. (\ref{eq:partial derivatives x of IoU}) and (\ref{eq:partial derivatives y of IoU}), we can see that the gradients for the position parameters \( x \) and \( y \) are determined by the change in the intersection area \( I \) with respect to \( x \) and \( y \). Specifically, the terms \( \frac{\partial I}{\partial x} \) and \( \frac{\partial I}{\partial y} \) represent how the intersection area changes as the predicted box moves. When the predicted box moves in a direction that increases \( I \), the gradients for \( x \) and \( y \) push the box further in that direction. Conversely, if moving the box reduces \( I \), the gradients reverse direction to encourage alignment. This mechanism ensures that the predicted box is pushed toward the ground truth box, maximizing the intersection area \( I \).

For the size parameters \( w \) and \( h \), the gradients in Eq. (\ref{eq:partial derivatives w of IoU}) and (\ref{eq:partial derivatives h of IoU}) consist of two terms. The first term, \( -\frac{S_g + S}{U^2} \cdot \frac{\partial I}{\partial w} \) (or \( -\frac{S_g + S}{U^2} \cdot \frac{\partial I}{\partial h} \)), encourages increasing the intersection area \( I \). If increasing \( w \) (or \( h \)) enlarges \( I \), the gradient pushes \( w \) (or \( h \)) to grow further. On the other hand, if increasing \( w \) (or \( h \)) reduces \( I \), the gradient reverses direction to shrink the box. The second term, \( \frac{I}{U^2} \cdot \frac{\partial S}{\partial w} \) (or \( \frac{I}{U^2} \cdot \frac{\partial S}{\partial h} \)), penalizes increasing the predicted box area \( S \). Since \( \frac{\partial S}{\partial w} = h \) and \( \frac{\partial S}{\partial h} = w \) are always positive, this term pushes \( w \) and \( h \) to decrease. Together, these two terms balance the objectives of \textbf{maximizing \( I \)} and \textbf{minimizing \( S \)}.

Overall, the $L_{IoU}$ achieves its optimization goals through the following mechanisms: (1) For the position parameters \( x \) and \( y \), the gradients align the predicted box with the ground truth, maximizing the intersection area \( I \); (2) For the size parameters \( w \) and \( h \), the gradients balance the competing objectives of increasing \( I \) and decreasing \( S \). This ensures that the predicted box tightly fits the ground truth, with a \textbf{large intersection area \( I \) and a small predicted area \( S \)}, thus maximizing the IoU value.

This internal optimization mechanism of \( L_{\text{IoU}} \) inherently avoids the \textit{Bbox Enlargement Problem}. Since both terms in InterpIoU are still based on IoU computations, this desirable property is inherited and continues to contribute to its effectiveness in improving Bbox localization.

\begin{table*}
  \centering
  \small
  \caption{Performance comparison of IoU-based loss functions using SSD and YOLOv8 on the VOC dataset. The columns for Bottle, Pottedplant, and Person show class-wise $AP_{50}$ results.}
  \label{tab:iou_loss_comparison}
  \begin{tabular}{l|l|ccc|ccc}
    \toprule
    Model & Method & Bottle & Pottedplant & Person & $AP_{50}$ & $AP_{75}$ & AP \\
    \midrule
    \multirow{7}{*}{SSD} 
    & GIoU       & 47.5 & 48.8 & 77.0 & 76.6 & 50.4 & 42.4 \\
    & CIoU       & 47.3 & 49.6 & 76.8 & 76.6 & 51.8 & 42.9 \\
    & DIoU       & 48.1 & 48.7 & 77.2 & 76.7 & 52.7 & 44.8 \\
    & SIoU       & 50.7 & 51.0 & 78.7 & 77.3 & 55.1 & 46.4 \\
    & PIoU       & 51.3 & 51.6 & 78.6 & 77.5 & 55.3 & 45.4 \\
    \cmidrule{2-8}
    & InterpIoU  & \textbf{52.0} & \textbf{53.0} & \textbf{79.2} & \textbf{77.7} & 56.0 & 45.6 \\
    & D-InterpIoU  & 50.2 & 51.0 & 78.8 & 77.6 & \textbf{56.3} & \textbf{47.0} \\
    \midrule
    \multirow{8}{*}{YOLOv8} 
    & GIoU       & 81.0 & 69.5 & 92.4 & 88.4 & 78.1 & 71.7 \\
    & CIoU       & 81.5 & 67.7 & 92.5 & 88.3 & 77.5 & 71.3 \\
    & DIoU       & 80.5 & 67.5 & 92.4 & 88.3 & 77.8 & 71.5 \\
    & SIoU       & 81.0 & 68.6 & 92.6 & 88.3 & 77.8 & 71.5 \\
    & PIoU       & 79.6 & 69.3 & 92.4 & 88.4 & 77.7 & 71.5 \\
    \cmidrule{2-8}
    & InterpIoU  & 81.3 & 68.2 & 92.4 & 88.6 & 77.6 & 71.6 \\
    & D-InterpIoU& \textbf{81.9} & \textbf{69.5} & \textbf{92.6} & \textbf{88.7} & \textbf{78.4} & \textbf{71.9} \\
    \bottomrule
  \end{tabular}
\end{table*}

\begin{table*}[ht]
    \centering
    \caption{Performance comparison of IoU-based loss functions using YOLOv8 on the VisDrone dataset. The columns for People, Car, Van and Truck show class-wise $AP_{50}$ results.}
    \label{tab: yolo_visdrone}
    \begin{tabular}{l|cccc|ccc}
      \toprule
      Method & People & Car & Van & Truck & $AP_{50}$ & $AP_{75}$ & AP \\
      \midrule
      GIoU    & 37.6 & 81.6 & 47.0 & 39.1 & 44.0 & 27.4 & 26.9 \\
      CIoU    & 37.5 & 81.7 & 46.9 & 42.6 & 44.1 & 27.8 & 27.0 \\
      DIoU    & 37.3 & 81.9 & 46.6 & 41.5 & 44.1 & 27.8 & 27.1 \\
      SIoU    & 38.1 & 81.7 & 46.7 & 41.4 & 44.2 & 27.6 & 27.0 \\
      PIoU    & 38.0 & 81.7 & 47.4 & 41.7 & 43.9 & 27.6 & 26.9 \\
      \midrule
      InterpIoU & 37.8 & 81.7 & 46.1 & 42.8 & 44.4 & 27.7 & 27.1 \\
      D-InterpIoU & \textbf{38.9} & \textbf{82.0} & \textbf{47.7} & \textbf{43.5} & \textbf{44.8} & \textbf{28.5} & \textbf{27.4} \\
      \bottomrule
    \end{tabular}
  \end{table*}
  \begin{table*}[ht]
      \centering
      \caption{Performance comparison of different IoU-based loss functions on the COCO dataset using YOLOv8m and DINO.}
      \label{tab: iou_loss_coco}
      \small 
      \begin{tabular}{l|l|ccc|ccc}
        \toprule
        Model & Method & m$AP$ & $AP_{50}$ & $AP_{75}$ & $AP_s$ & $AP_m$ & $AP_l$ \\
        \midrule
        \multirow{7}{*}{YOLOv8}
        & GIoU       & 49.1 & 66.0 & 53.3 & 30.3 & 54.5 & 65.3 \\
        & CIoU       & 49.2 & 66.0 & 53.6 & 30.6 & 54.6 & 65.5 \\
        & DIoU       & 49.1 & 65.9 & 53.4 & 30.2 & 54.4 & 65.7 \\
        & SIoU       & 49.3 & 66.3 & 53.5 & 30.2 & 54.6 & \textbf{66.1} \\
        & PIoU       & 49.1 & 66.1 & 53.4 & 30.6 & 54.6 & 65.1 \\
        \cmidrule{2-8}
        & InterpIoU  & \textbf{49.4} & \textbf{66.4} & 53.7 & 30.7 & \textbf{54.6} & 65.6 \\
        & D-InterpIoU  & 49.3 & 66.2 & \textbf{53.7} & \textbf{31.5} & 54.5 & 65.6 \\
        \midrule
        \multirow{7}{*}{DINO}
        & GIoU       & 49.2 & 66.7 & 53.7 & 32.3 & 52.2 & 63.6 \\
        & CIoU       & 47.5 & 64.6 & 51.7 & 30.1 & 50.7 & 61.2 \\
        & DIoU       & 49.4 & 67.1 & 53.9 & 32.3 & 53.0 & 63.9 \\
        & SIoU       & 49.4 & 67.0 & 54.2 & 32.3 & 52.6 & 64.0 \\
        & PIoU       & 49.6 & 67.2 & 54.0 & 31.9 & 53.0 & 63.7 \\
        \cmidrule{2-8}
        & InterpIoU  & \textbf{49.7} & \textbf{67.3} & 54.0 & \textbf{33.1} & 52.5 & \textbf{64.3} \\
        & D-InterpIoU & 49.4 & 66.9 & \textbf{54.2} & 32.9 & \textbf{53.1} & 63.9 \\
        \bottomrule
      \end{tabular}
    \end{table*}

\subsection{Enhancing Small Target Detection}

Detecting small objects remains a difficult problem in object detection because they occupy only a small part of the image and contain limited visual information. In the COCO dataset~\cite{coco}, objects smaller than $32 \times 32$ pixels are considered small, which usually means they take up less than $1/400$ of the image size when resized to $640 \times 640$ for training. In remote sensing datasets like VisDrone~\cite{VisDrone}, small targets such as vehicles and pedestrians often cover only a few dozen pixels.

Many previous IoU-based losses try to improve performance by adding penalty terms based on geometric features. However, these penalty terms are often sensitive to the size, shape, and distribution of bounding boxes, and their regression performance can drop significantly when handling small objects. Our proposed InterpIoU avoids this problem by using interpolated boxes between the prediction and ground truth to guide optimization. This approach does not rely on specific geometric structures and provides more stable gradients in various cases.

To show this, we designed simulation experiments where the ground truth area $S_{gt}$ was set to $1/32$, $1/64$, and $1/128$ of the image area. As shown in Fig.~\ref{fig:loss-performance-loss-cirteria}, InterpIoU shows more stable and faster convergence than traditional IoU-based losses when the target becomes smaller, showing its advantage in small object detection.

\section{Experiments and Analysis}
\label{sec: experiments section}

\subsection{Experimental Setup}

\textbf{Datasets.}  
To comprehensively evaluate the effectiveness of the proposed InterpIoU, we conduct experiments on three widely used benchmarks.  
(1) \textbf{Pascal VOC 2007} \cite{voc} provides a relatively simpler environment, allowing clearer comparisons of baseline performance.  
(2) \textbf{VisDrone} \cite{VisDrone} focuses on drone-based visual data, which contains numerous urban scenes with dense object distributions and consistently small targets—making it ideal for evaluating the small-object detection capability of our method. 
(3) \textbf{MS COCO} \cite{coco} is used to test performance in complex, real-world scenarios with diverse object categories and scales.  

\textbf{Models.}  
To ensure fair and comprehensive evaluation across different model architectures, we adopt three representative detectors:  
(1) \textbf{SSD} \cite{ssd}, a classical one-stage object detector widely used as a baseline. Its simple architecture and lack of aggressive training tricks make it ideal for isolating the effect of the loss function, enabling a clear assessment of localization quality.  
(2) \textbf{YOLOv8-m} \cite{yolov8}, a modern real-time detector that incorporates advanced design components and training heuristics. Improvements on such a strong baseline highlight the compatibility and robustness of our method in modern detection pipelines. We select the $m$ scale to balance performance and training cost; unless otherwise specified, YOLOv8 refers to this setting throughout the paper.  
(3) \textbf{DINO-4scale} \cite{dino}, a powerful transformer-based detector with strong performance on COCO. It allows us to test the generalization of our loss beyond anchor-based frameworks.  

\textbf{Training Settings.}
For DINO, we used the official training scheme with batchsizes at 8 under 12 epochs. The YOLOv8 is also trained under Ultralytics \cite{yolov8} framework, with COCO using 500 epoches and batchsize at 4, both VisDrone and VOC using batchsize 16 and 150 epochs. SSD is trained using batchsize 32 with 140k iterations. The Interpolation Coefficient and the upper and lower bound for dynamic alpha will be specified in the following statement.

  \begin{figure}[h!tb]
    \centering
    \includegraphics[width=\linewidth]{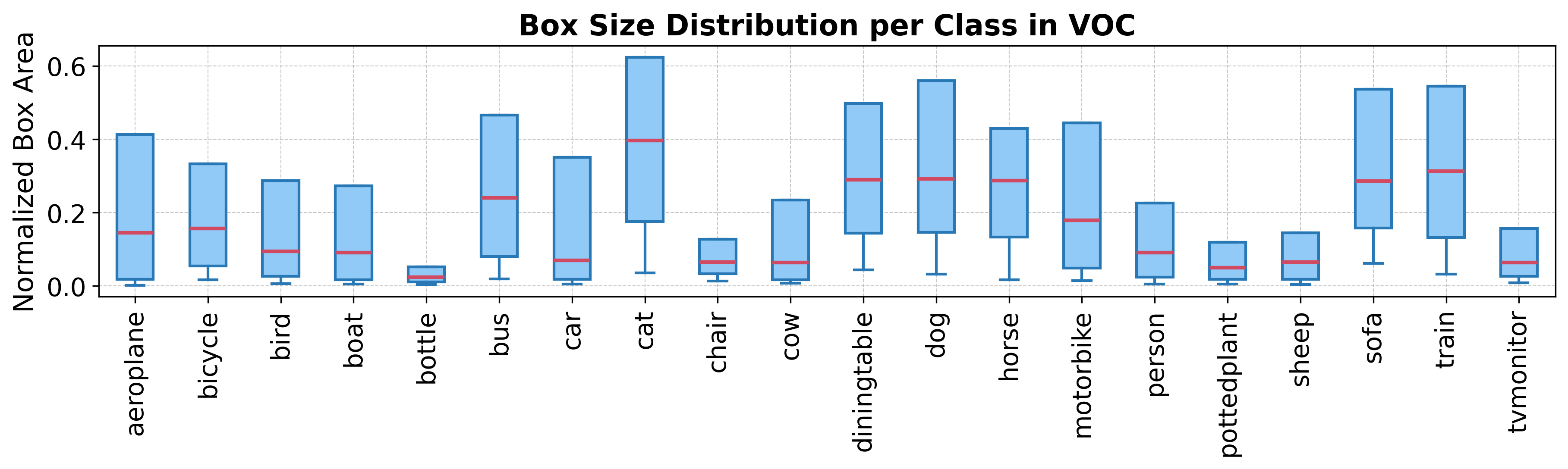}
    \caption{Comparison of relative Bbox area distributions across object classes in the VOC dataset (test split only). Each box represents the distribution of \textit{width~$\times$~height} (normalized by image size) for all ground truth boxes in a given class. Outlier values are omitted for clarity. }
    \label{fig:voc-box-sizes}
\end{figure}

\subsection{Comparison Experiments}
\label{sec: comparision experiments}

\textbf{Experiments on PASCAL VOC.}  
We begin our evaluation on the PASCAL VOC dataset, which features relatively clean scenes and moderate object sizes. As summarized in Table~\ref{tab:iou_loss_comparison}, both InterpIoU and its dynamic variant, D-InterpIoU, consistently outperform prior IoU-based loss functions on this dataset, across both SSD and YOLOv8. InterpIoU uses a fixed interpolation coefficient of $0.98$, while D-InterpIoU dynamically adjusts $\alpha$ within the range $[0.95, 0.99]$.

In addition to overall AP scores, we also highlight class-wise results for three representative small-object categories: \textit{bottle}, \textit{pottedplant}, and \textit{person}, selected based on their small normalized box sizes as shown in Fig.~\ref{fig:voc-box-sizes}. On these challenging classes, both InterpIoU and D-InterpIoU consistently outperform all other baselines, demonstrating their robustness in fine-grained localization tasks.


\textbf{Experiments on VisDrone.}  
We further evaluate our method on the VisDrone dataset, which features dense, small-scale objects captured from drone perspectives. As shown in Table~\ref{tab: yolo_visdrone}, both InterpIoU and D-InterpIoU outperform all existing IoU-based losses across most metrics. InterpIoU is configured with a fixed interpolation coefficient of $0.98$, while D-InterpIoU dynamically adjusts $\alpha$ within the range $[0.60, 0.99]$.

Notably, D-InterpIoU achieves the highest scores on $AP_{50}$, $AP_{75}$, and overall AP, demonstrating its robustness in localizing small objects under cluttered real-world scenes. Analyzing the ground-truth bounding boxes, we find that the average normalized area is approximately 0.002 (i.e., 0.2\% of image size), meaning that most targets fall well within the small-object category. This makes VisDrone an ideal benchmark for evaluating fine-grained localization, and our method’s consistent gains in this setting validate its strength in handling small-scale instances. In particular, we highlight D-InterpIoU’s improvements in four representative categories—\textit{people}, \textit{car}, \textit{van}, and \textit{truck}—where objects are typically small and densely distributed.

\textbf{Experiments on MS COCO.}  
Finally, we evaluate our methods on the MS COCO dataset, which features a wide variety of object categories, scales, and cluttered backgrounds—posing a strong challenge for object localization. InterpIoU is applied with a fixed interpolation coefficient $\alpha=0.98$, while D-InterpIoU dynamically adjusts $\alpha$ in the range $[0.90, 0.99]$.

As shown in Table~\ref{tab: iou_loss_coco}, both InterpIoU and D-InterpIoU achieve top-tier results across different detectors. On YOLOv8, InterpIoU attains the highest $AP_{50}$ and overall mAP, while D-InterpIoU matches the best $AP_{75}$ and delivers the strongest $AP_s$ (+0.8 over the best baseline), confirming its advantage in localizing small-scale objects even in complex scenes. On the transformer-based DINO model, InterpIoU again provides the highest mAP and $AP_s$, while D-InterpIoU shows competitive results with leading $AP_{75}$ and the best $AP_m$. These consistent gains across scales and metrics highlight the robustness and versatility of our proposed interpolation-based loss family.

In contrast, D-InterpIoU does not show further improvement on COCO and slightly underperforms compared to its static counterpart. This may be attributed to the highly varied object sizes and complex scene layouts in COCO, where a dynamic interpolation coefficient driven by the IoU value might introduce unstable or overly conservative adjustments. In such settings, a fixed but well-chosen coefficient (e.g., \(\alpha = 0.98\)) offers more stable gradient behavior during training. Nevertheless, as shown in other experiments (e.g., on VisDrone and VOC), D-InterpIoU still demonstrates advantages under consistent object scale distributions, confirming its potential under appropriate data conditions.

\begin{figure*}[h!tb]
  \centering 
  \includegraphics[scale=0.60]{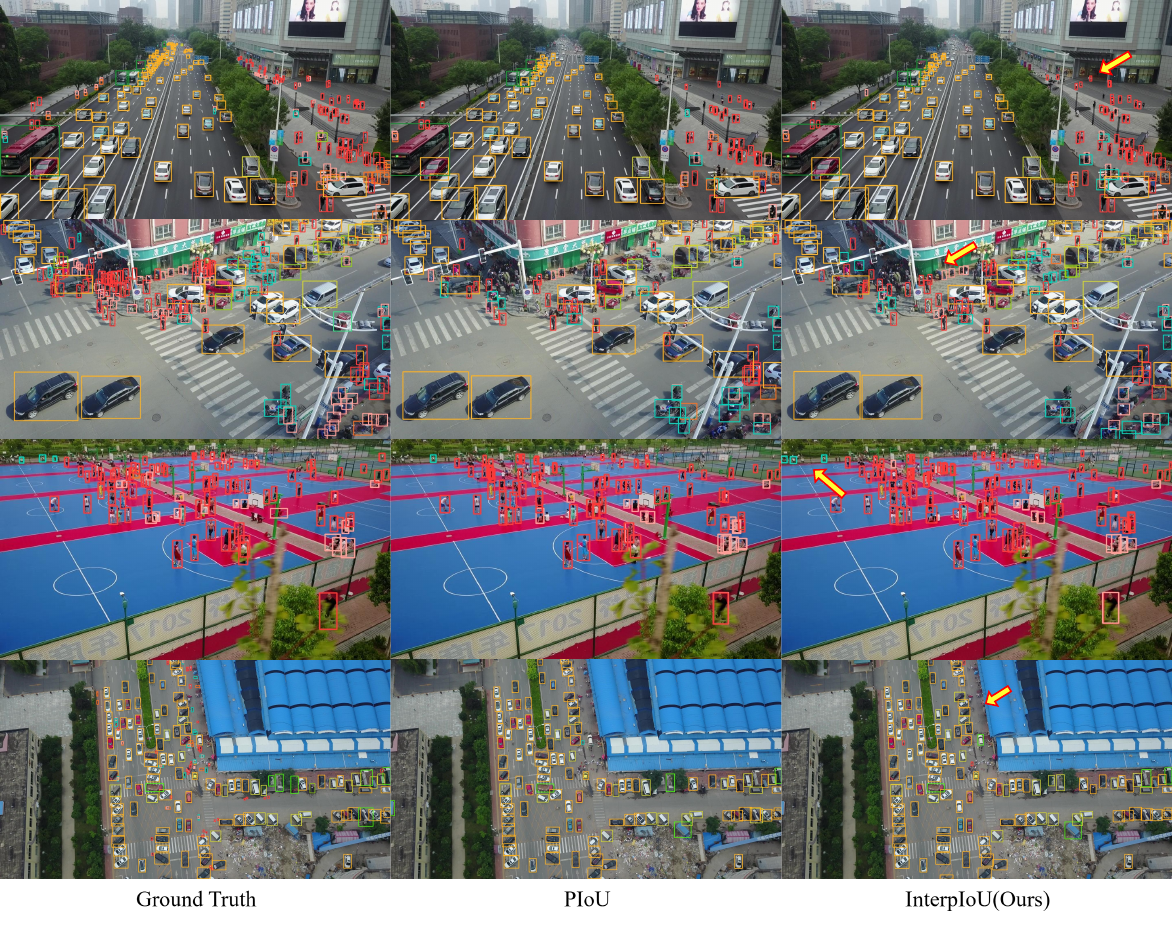}
  \caption{Visualization results on the VisDrone dataset. The first column shows the Ground Truth (GT), the second column shows the results from the PIoU method, and the third column shows the results from our proposed InterpIoU method. Our IoU loss effectively improves the detection of small objects, demonstrating better localization and accuracy compared to the baseline.}
  \label{fig:yolov8_visdrone_visualization}
\end{figure*}

\subsection{Ablation Study on the Interpolation Coefficient}
\label{sec: abla on alpha}
As discussed in Sec.~\ref{sec: interpiou proposal}, the interpolation coefficient \( \alpha \) plays a crucial role in ensuring that the interpolated boxes sufficiently overlap with the target boxes, especially when the predicted and ground truth boxes are far apart. To investigate its impact, we conduct an ablation study on both the VisDrone and VOC datasets using YOLOv8s, trained for 100 epoches, with results summarized in Table~\ref{tab:alpha_ablation}.

We observe that performance is indeed sensitive to the choice of \( \alpha \). On the VisDrone dataset, the best mAP is achieved at \( \alpha = 0.98 \), whereas on VOC, the best result occurs near \( \alpha = 0.95 \). Notably, in the lower range (\( \alpha \in [0.50, 0.90] \)), there is a clear upward trend in accuracy as \( \alpha \) increases, indicating that smaller values lead to weaker interpolated gradients and less effective supervision. This trend reflects the importance of sufficient overlap between interpolated and target boxes. 

Once \( \alpha \) reaches around 0.90, performance enters a relatively stable plateau (\( \alpha \in [0.90, 0.99] \)), suggesting that as long as sufficient interpolation is ensured, the exact value of \( \alpha \) becomes less critical. 

Notably, the dynamic version (\(clamp=[0.00, 0.99]\)) achieves competitive or superior performance compared to all static settings across most metrics and datasets. This supports our hypothesis that adjusting \( \alpha \) dynamically based on the IoU value enhances adaptability to different regression scenarios, improving performance on both challenging small-object datasets (VisDrone) and high-quality annotation datasets (VOC). In particular, the dynamic approach achieves the highest $AP_{75}$ on VisDrone and the highest mAP on VOC, validating its generalization capability across detector types and data domains.

\begin{table}
  \centering
  \caption{Ablation Experiments of YOLOv8s on VOC and VisDrone.}
  \label{tab:alpha_ablation}
  \begin{tabular}{l|ccc|ccc}
    \toprule
    \multirow{2}{*}{$\alpha$} & \multicolumn{3}{c}{VisDrone} & \multicolumn{3}{c}{VOC} \\
    \cmidrule(r){2-4} \cmidrule(r){5-7}
    & $AP_{50}$ & $AP_{75}$ & m$AP$ & $AP_{50}$ & $AP_{75}$ & m$AP$ \\
    \midrule
    dyn.    & 40.6  & \textbf{25.1} & \textbf{24.6} & 86.5 & 74.6 & \textbf{68.0}  \\
    0.99    & 40.6 & 24.8 & 24.3 & 86.4 & 74.1 & 67.5  \\
    0.98    & \textbf{40.7} & 24.7 & 24.4 & 86.5 & 74.2 & 67.7\\
    0.97    & 40.2 & 24.4 & 24.2 & 86.5 & 74.5 & 67.8 \\
    0.95    & 40.6 & 24.3 & 24.3 & 86.5 & 74.4 & 67.9 \\
    0.90    & 40.2 & 24.7 & 24.3 & \textbf{86.7} & 74.6 & 67.8 \\
    0.80    & 40.2 & 25.1 & 24.4 & 86.3 & \textbf{74.7} & 67.9 \\
    0.70    & 40.5 & 24.8 & 24.4 & 86.1 & 74.5 & 67.8 \\
    0.60    & 40.3 & 25.0 & 24.4 & 86.0 & 74.3 & 67.7 \\
    0.50    & 39.6 & 24.6 & 24.3 & 85.7 & 74.7 & 67.8  \\
    \bottomrule
  \end{tabular}
\end{table}

\subsection{Visualization}
\label{sec: visualization}

Fig.~\ref{fig:yolov8_visdrone_visualization} presents a qualitative analysis on the VisDrone dataset to intuitively demonstrate the performance improvements brought by our proposed InterpIoU. Using YOLOv8m as the detector, we compare our method against the baseline PIoU and Ground Truth (GT) across four distinct and challenging aerial scenes.

These visualizations highlight the strength of our method in detecting small and difficult objects. In the first row, which features strong perspective distortion, InterpIoU detects more distant vehicles and small pedestrians along the roadside, outperforming the baseline in both precision and recall. This robustness is further evident in the fourth row's top-down view, where our approach confidently identifies the object indicated by the arrow, suggesting that the well-optimized BBR effectively enhances the model's ability to mitigate contextual interference.

InterpIoU also shows clear advantages in densely crowded scenes. In the second row, depicting a crowded street intersection, the baseline fails to detect several occluded individuals, while our method recovers more of them despite high occlusion. The third row confirms this benefit again, where a missed target in a crowded area is correctly identified by our approach. This suggests that the stable optimization afforded by our interpolation strategy helps distinguish targets from both dense crowds and distracting backgrounds.

Together, these qualitative findings visually confirm the benefits of InterpIoU in real-world scenarios. They support our quantitative results, demonstrating that interpolation-based loss design not only enhances gradient behavior theoretically but also leads to tangible improvements in detection accuracy and robustness.

\section{Conclusion}
\label{sec: conclusion}

In this paper, we introduced InterpIoU, a novel IoU-based loss function that replaces handcrafted geometric penalties with a principled penalty term defined by the IoU between an interpolated box and the ground truth. This formulation preserves the desirable optimization behavior of standard IoU loss while addressing its gradient vanishing issue in non-overlapping cases. By eliminating the need for manually designed geometric terms, InterpIoU also inherently avoids undesired effects such as bounding box enlargement and demonstrates superior robustness in small object localization.

Furthermore, we proposed Dynamic InterpIoU, which adjusts the interpolation coefficient dynamically based on the IoU value, enabling better adaptation to objects with diverse sizes and spatial distributions. Through both simulation analysis and extensive experiments on COCO, VisDrone, and PASCAL VOC, we demonstrated that our methods consistently outperform state-of-the-art IoU-based losses across various detection frameworks. In particular, the improvements in small object detection confirm the effectiveness and general applicability of our approach.

{
    \small
    \bibliographystyle{ieeenat_fullname}
    \bibliography{interpiou_arxiv}
}


\end{document}